\documentclass[10pt, conference, compsocconf]{IEEEtran}

\usepackage{cite}
\usepackage{amsmath,amssymb,amsfonts}
\usepackage{algorithmic}
\usepackage{graphicx}
\usepackage{textcomp}
\usepackage{xcolor}
\usepackage{subfig}
\usepackage{balance}

\begin{document}
\title{ Multiple Classifiers Based Adversarial Training for Unsupervised Domain Adaptation}

\author{\IEEEauthorblockN{Yiju Yang, Taejoon Kim}
\IEEEauthorblockA{Department of Electrical Engineering \& Computer Science\\ University of Kansas, Lawrence KS, USA, 66045\\
\{y150y133, taejoonkim\}@ku.edu}
\and
\IEEEauthorblockN{Guanghui Wang}
\IEEEauthorblockA{Department of Computer Science\\ Ryerson University, Toronto ON, Canada, M5B 2K3\\
wangcs@ryerson.ca}
}


	
\maketitle
\thispagestyle{empty}
\pagestyle{empty}

\begin{abstract}

Adversarial training based on the maximum classifier discrepancy between two classifier structures has achieved great success in unsupervised domain adaptation tasks for image classification. The approach adopts the structure of two classifiers, though simple and intuitive, the learned classification boundary may not well represent the data property in the new domain. In this paper, we propose to extend the structure to multiple classifiers to further boost its performance. To this end, we develop a very straightforward approach to adding more classifiers. We employ the principle that the classifiers are different from each other to construct a discrepancy loss function for multiple classifiers. The proposed construction method of loss function makes it possible to add any number of classifiers to the original framework. The proposed approach is validated through extensive experimental evaluations. We demonstrate that, on average, adopting the structure of three classifiers normally yields the best performance as a trade-off between accuracy and efficiency. With minimum extra computational costs, the proposed approach can significantly improve the performance of the original algorithm.
The source code of the proposed approach can be downloaded from  \url{https://github.com/rucv/MMCD\_DA}.
\end{abstract}

\vspace{12pt}
\section{INTRODUCTION}
Thanks to the powerful capability of convolutional neural networks (CNN) in extracting image features, CNN-based deep learning models have achieved significant breakthroughs in various computer vision applications like classification \cite{cen2021deep, chen2021few, li2021sgnet}, segmentation \cite{he2021sosd, patel2021enhanced}, detection \cite{ma2020mdfn, xu2020adaptively}, recognition \cite{gajurel2021fine, sajid2021audio}, tracking \cite{zhang2020efficient}, image generation \cite{xu2021domain}, and counting \cite{sajid2021towards}. However, most current neural network models rely heavily on accurately labeled data. When the trained models are applied to unfamiliar scenes, the performance of these deep learning models will be degraded significantly. To address this issue, one commonly used approach is based on transfer learning to fine-tune the existing models using the labeled data in the new environment by fixing the parameters of the first few layers of the network and using the new data to fine-tune the parameters of the last few layers of the network.
The approach is widely adopted since the amount of labeled data it needs is much less than that of supervised learning and it is easy to implement. However, the performance of this approach is normally inferior to fully supervised learning. In practice, data collection is much easier than data annotation. Thus, how to use the unlabeled data for unsupervised learning has attracted a lot of attention in recent years.

In order to solve this problem using the unlabeled data, many researchers try to make deep learning models to learn the common features (i.e., domain invariant features) in different scenarios. If the models can discover common features among different scenes, they are able to effectively classify these scenes. In the study \cite{ganin2015unsupervised}, RevGrad was proposed to use domain classifiers to perform image classification through unsupervised domain adaptation. This method can effectively merge the two domains and allow the feature extractor to extract the domain-invariant features.
Inspired by this study, more and more researchers have begun to explore how to use deep-learning models to extract the domain invariant features, which is one of the main challenges in unsupervised domain adaptation.

Recently, Maximum Classifier Discrepancy (MCD) \cite{saito2018maximum} was proposed using an adversarial training framework. In this framework, the two classifiers are trained by using the maximum classifier discrepancy, and the feature extractor parameters are adjusted inversely through the decision boundary of the two classifiers so that the source domain and the target domain are fused together.
In MCD, only two classifiers are used which is straightforward, however, the classification boundary learned by only two classifiers may not be very discriminative. Is it possible to employ more classifiers and what performance benefits can be achieved? The paper will investigate this problem and study the influence of using multiple classifiers. We propose an effective and convenient approach to add and train multiple classifiers under the MCD framework. We also compare the performance of approaches using difference number of classifiers.

The main contributions of this paper include:

\begin{enumerate} 
\item We propose a new approach to construct a multi-difference loss function based on the principle of different classifiers, making it possible to add any number of classifiers into the structure.
\item We empirically prove that multi-classifier structures have obvious performance advantages in comparison to the 2-classifier structure, however, adding the number of classifiers could not increase its performance indefinitely.
\item Based on our experimental study, we recommend the structure of using three classifiers as a trade-off between the classification accuracy and the computational complexity. Using three classifiers outperforms the 2-classifier counterpart by a large margin.
\end{enumerate}

\section{Related Work}

 Domain adaptation is a widely used technique in many computer vision tasks to improve the generalization ability of a model trained on a single domain. In this section, we describe some existing domain adaptation methods.

 {\bf Learn domain invariant features.}
  Recently, Chen et al. \cite{chen2020simple} explored what enables the contrastive prediction tasks to learn useful representations.
 \cite{Carlucci_2019_CVPR} proposed to learn the semantic labels in a supervised fashion, and broadens its understanding of the data by learning from self-supervised signals how to solve a jigsaw puzzle on the same images. Yang et al. \cite{yang2021unsupervised} proposed a  dual-module network to learn more domain invariant features.

 {\bf Distance-based methods.}
 Aligning the distribution between the source domain and the target domain is a very common method in solving unsupervised domain adaptation problems. Maximum Mean discrepancy (MMD) is a method of measuring the difference between two distributions \cite{gretton2012kernel,long2017deep,tzeng2014deep,long2015learning}. 
 DAN \cite{long2015learning} explored the multi-core version of MMD to define the distance between two distributions. 
 JAN  \cite{long2017deep} learned a transfer network by aligning the joint distributions of multiple domain-specific layers across the domains based on a joint maximum mean discrepancy (JMMD) criterion.
  \cite{long2016unsupervised} enabled the classifier adaptation by plugging several layers into the deep network to explicitly learn the residual function with reference to the target classifier.
 CMD \cite{zellinger2017central} defined a metric on the set of probability distributions on a compact interval.
 
 To solve the problem of unbalanced datasets, Deep Asymmetric Transfer Network (DATN)  \cite{wang2018deep} proposed to learn a transfer function from the target domain to the source domain and meanwhile adapting the source domain classifier with more discriminative power to the target domain.
 DeepCORAL \cite{sun2016deep} proposed to build a specific deep neural network by aligning the distribution of second-order statistics to limit the invariant domain of the top layer. 
  \cite{chen2020homm} proposed a Higher-order Moment Matching (HoMM) method to minimize the domain discrepancy.

 {\bf Adversarial methods.}
 Adversarial training is another very effective method to transfer domain information. 
 Inspired by the work of gradient reversal layer  \cite{ganin2015unsupervised}, a group of domain adaptation methods has been proposed based on adversarial learning. 
RevGrad  \cite{ganin2015unsupervised} proposed to learn the global invariant feature by using a discriminator that is used to reduce the discriminative features in the domain.
 Deep Reconstruction-Classification Networks (DRCN)  \cite{ghifary2016deep} jointly learned a shared encoding representation for supervised classification of the labeled source data, and unsupervised reconstruction of the unlabeled target data.
  \cite{bousmalis2016domain} extracted image representations that are partitioned into two subspaces. Adversarial Discriminative Domain Adaptation (ADDA)  \cite{tzeng2017adversarial} trained two feature extractors for the source and target domains respectively, to generate embeddings to fool the discriminator.
  
 Maximum Classifier Discrepancy (MCD) \cite{saito2018maximum} was proposed by exploring task-specific decision boundaries. 
 CyCADA \cite{hoffman2018cycada} introduced a cycle-consistency loss to match the pixel-level distribution. 
 SimeNet  \cite{pinheiro2018unsupervised} solved this problem by learning the domain invariant features and the categorical prototype representations.
 CAN  \cite{kang2019contrastive} optimized the network by considering the discrepancy between the intra-class domain and the inter-class domain.
 Graph Convolutional Adversarial Network (GCAN)  \cite{ma2019gcan} realized the unsupervised domain adaptation by jointly modeling data structure, domain label, and class label in a unified deep model.
  \cite{gong2019dlow} proposed a domain flow generation (DLOW) model to bridge two different domains by generating a continuous sequence of intermediate domains flowing from one domain to the other.
  \cite{cai2019learning} employed a variational auto-encoder to reconstruct the semantic latent variables and domain latent variables behind the data.
 Drop to Adapt (DTA)  \cite{lee2019drop} leveraged adversarial dropout to learn strongly discriminative features by enforcing the cluster assumption.
 Instead of representing the classifier as a weight vector,  \cite{lu2020stochastic} modeled it as a Gaussian distribution with its variance representing the inter-classifier discrepancy.

\section{Proposed Method}
 In this work, we consider the close-set an unsupervised domain adaptation problem \cite{yang2021unsupervised}.
 Suppose we have a source domain $D_s = {\{(X_s, Y_s)\}} = {\{(x_s^i, y_s^i)\}}_{i=1}^{n_s}$ with $n_s$ labeled samples and a target domain $D_t = {\{(X_t)\}} = {\{(x_t^i)\}}_{i=1}^{n_t}$ with $n_t$ unlabeled samples.
 The two domains share the same label space $Y = \{1,2,3,...,K\}$, where $K$ is the number of categories. We assume that the source sample $x_s$ belongs to the source distribution $P_s$, and the target sample $x_t$ belongs to the target distribution $P_t$, where $P_s \neq P_t$. 
 Our goal is to train a classifier $f_\theta (x)$ that can minimize the target risk $\epsilon _t = E_{x\in D_t}[f_\theta (x) \neq y_t]$, where $f_\theta (x)$ represents the output of the deep neural network, and $\theta$ represents the model parameters to be learned.

\subsection{Discrepancy Loss for Two Classifiers}
 We follow the discrepancy loss in  \cite{saito2018maximum} and use the absolute value of the difference between the probability outputs of the two classifiers as the discrepancy loss: 
 \begin{equation} 
 L_2(p^1,p^2) = \frac{1}{K} \sum^K_{k=1} |p^1_k - p^2_k| \end{equation}
 where $p^1$ and $p^2$ are the probability outputs of the two classifiers respectively, which are the prediction scores for all the categories, $K$ is the number of categories, and $p^1_k$ and $p^2_k$ are the specific values of their $k$-{th} category.

\subsection{Discrepancy Loss for Multiple Classifiers}
 We construct the discrepancy loss of the multiple classifiers {$L_m()$} based on the principle that the classifiers are different from each other. 
 In our training step, we need to maximize the difference between the classifiers. If the classifiers cannot maintain the principle of mutual difference, then there will be a situation where the overall discrepancy is maximized but some local discrepancy is close to zero. Some classifier parameters will become the same in this process. 
 If the parameters of multiple classifiers tend to be the same when training multiple classifiers, it will lead to a collapse of the model training. This conflict will make the performance of the model even worse.

 The following three discrepancy losses are designed for the cases of 3 classifiers, 4 classifiers, and $n$ classifiers.
 
 {\bf For 3 classifiers:}
 \begin{equation*}L_3(p^1,p^2,p^3)
 = L_2(p^1,p^2)+L_2(p^1,p^3)+L_2(p^2,p^3) \end{equation*}
 
 {\bf For 4 classifiers:}
 \begin{equation*}
 \begin{aligned}
 L_4(p^1,p^2,p^3,p^4) = L_2(p^1,p^2) + L_2(p^1,p^3) + L_2(p^1,p^4)\\
 + L_2(p^2,p^3) + L_2(p^2,p^4) \\
 + L_2(p^3,p^4) \end{aligned}
 \end{equation*}
 
 {\bf For $n$ classifiers:}
  \begin{equation*}
  \begin{aligned}L_n(p^1,\ldots,p^n) \!=\! L_2(p^1,p^2) \!+\! L_2(p^1,p^3) + \cdots + L_2(p^1,p^n) \\
  + L_2(p^2,p^3) + \cdots + L_2(p^2,p^n)\\
  + \cdots + \cdots\\
  + L_2(p^{n-1},p^n) \end{aligned}
  \end{equation*} 
 where $n$ is the number of classifiers, and $p^1$, $p^2$,$\ldots$, $p^n$ are the probability outputs of the $n$ classifiers respectively.

\subsection{Training Steps}
The proposed approach is based on the framework of two classifiers. In order to handle multiple classifiers, we replace the original discrepancy loss with the multi-classifier discrepancy loss under the original framework.
 
 {\bf Step 1:}
 We first directly employ the source domain data to train the model for one time so that the model can initially learn some source domain information (features).
 After this step, the decision boundary from each classifier can classify the features extracted from the source domain. 
 
  \begin{equation}
  \begin{aligned}
  Loss_1 =  L_{C1}(f_{\theta 1}(X_s),Y_s) + L_{C2}(f_{\theta 2}(X_s),Y_s) +  \cdots \\ 
  + L_{Cn}(f_{\theta n}(X_s),Y_s)
  \end{aligned}
  \end{equation}
 where $L_{C1}()$, $L_{C2}()$, $\ldots$, $L_{Cn}()$ denote the cross-entropy loss of the $1$-st, $2$-nd, and $n$-th classifier, respectively

 {\bf Step 2:}
 In this step, we use both the source and target domain data to train our model. During the training, we fix the parameters of the feature extractor and only adjust the parameters of the classifiers. When train using the source domain data, we follow the same routine as in the first step. 
 When training using the target domain data, we could not perform supervised training since the data have no labels. We employ the discrepancy loss for the target domain training at this step.  
 We employ the $n$-th classifier discrepancy loss to maximize the discrepancy between the results of the classifiers predicting the target domain. We maximize the discrepancy by reversing the gradient.
 
 The loss of the joint training process is defined as 
   \begin{equation}
  \begin{aligned}
  Loss_2 = Loss_s - Loss_t
  \end{aligned}
  \end{equation}
where  
  \begin{equation}
  \begin{aligned}
  Loss_s =  L_{C1}(f_{\theta 1}(X_s),Y_s) + L_{C2}(f_{\theta 2}(X_s),Y_s)  \\ 
  + \cdots + L_{Cn}(f_{\theta n}(X_s),Y_s)
  \end{aligned}
  \end{equation}
 is the loss for the source domain, and 
  \begin{equation}
  \begin{aligned}
  Loss_t = L_n(f_{\theta 1}(X_t),\ldots,f_{\theta n}(X_t))
  \end{aligned}
  \end{equation}
 is the loss for the target domain.

 {\bf Step 3:}
 In this step, we minimize the discrepancy in the prediction results of the features of the target domain extracted by the feature extractor.
 During this process, we fix the $n$-th classifier parameters and update the CNN’s parameters by back-propagation. Since the parameters of the classifiers are fixed, in order to minimize the prediction discrepancy of the $n$-th classifier, the CNN must adjust the parameters so that the features extracted from the target domain are consistent with the features obtained from the source domain, and as a result, they could achieve similar prediction results with totally different parameters.
  
  \begin{equation}
  \begin{aligned}
  Loss_3 = L_n(f_{\theta 1}(X_t),...,f_{\theta n}(X_t))
  \end{aligned}
  \end{equation}
 
 The source domain is trained by the labeled data, even when the classifier has the maximum discrepancy to the target domain, thus, it will not have much impact when detecting the source domain.

\section{Experimental Results}

\subsection{Experiments on the Toy Dataset}
\begin{figure}[t]
	\begin{minipage}[b]{0.32\columnwidth}
		\begin{center}
			\centerline{\includegraphics[width=0.9\columnwidth]{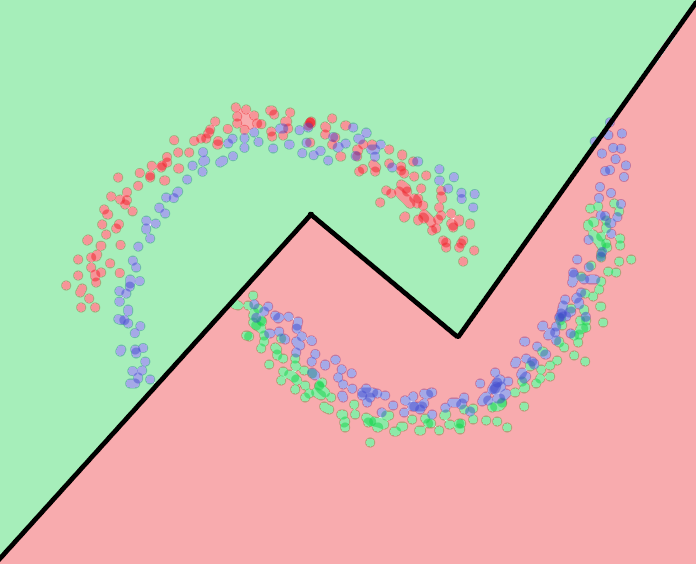}}
			{\footnotesize{a) Source Only}}
		\end{center}
	\end{minipage}
	\begin{minipage}[b]{0.32\columnwidth}
		\begin{center}
			\centerline{\includegraphics[width=0.9\columnwidth]{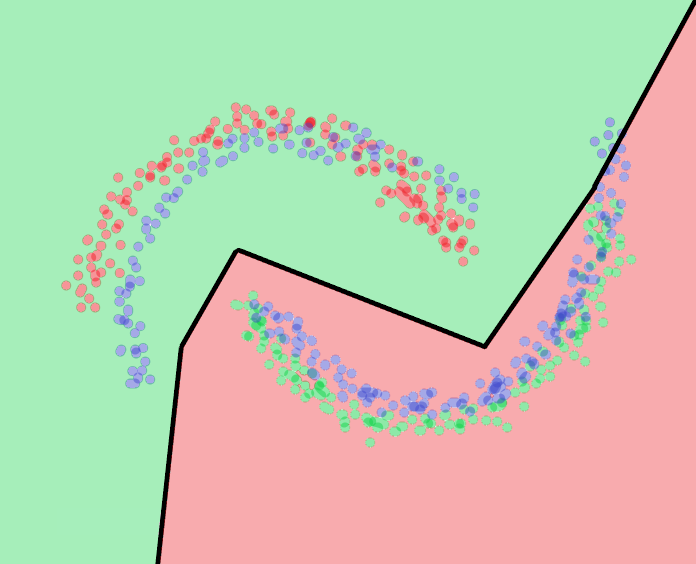}}
			{\footnotesize{b) Adapted: MCD}}
		\end{center}
	\end{minipage}
	\begin{minipage}[b]{0.32\columnwidth}
		\begin{center}
			\centerline{\includegraphics[width=0.9\columnwidth]{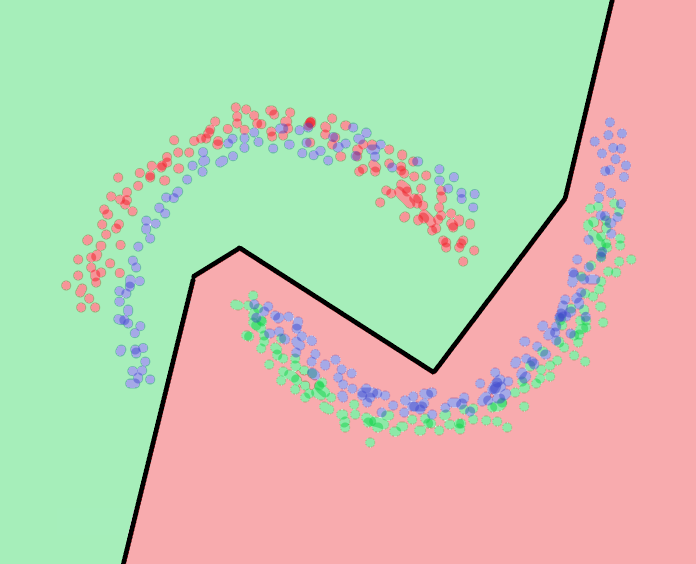}}
		    {\footnotesize{c) Adapted: Ours}}
		\end{center}
	\end{minipage}
	\caption{Experiments on the toy dataset. The red points and the green points are the two classes of the source domain data, while the blue points denote the target domain data. The dividing line in the middle is the decision boundary. We can see that the result using multiple classifiers is more effective than others.
	a) is the model trained only on the source data; 
	b) is the original { MCD \cite{lu2020stochastic}} with the two-classifier structure; and
    c) is the proposed approach with a three-classifier structure.}

	\label{fig:4}
\end{figure}

 In this experiment, we compared our proposed method with MCD and the source-only method. The purpose is to compare the decision boundaries of different approaches under the same task. 
 We followed the same experimental setting as \cite{yang2021unsupervised}\cite{lu2020stochastic}.
 We generated 300 source and target domain data and only gave labels to the source domain data.
 As shown in Fig \ref{fig:4}, we can see that the decision boundary under multiple classifiers has better unsupervised classification capability than MCD with a two-classifier structure and the source-only method.
 The decision boundaries are drawn considering both the source and target samples. The outputs of three or multiple classifiers make nearly the same prediction for the target samples, and they could classify most target samples correctly, so we randomly pick one of them for the illustration in Fig \ref{fig:4}.
 
\subsection{Experiments on Digits and Signed Datasets}\label{IV.B}
In this section, we evaluate our model using the following five datasets: MNIST \cite{lecun-mnisthandwrittendigit-2010}, Street View House Numbers (SVHN) \cite{netzer2011reading}, USPS  \cite{291440}, Synthetic Traffic Signs (SYN SIGNS) \cite{978-3-319-02895-8_52}, and the German Traffic Signs Recognition Benchmark (GTSRB) \cite{Stallkamp-IJCNN-2011}.

\textbf{MNIST:} The dataset contains images of digits 0 to 9 in different styles. It is composed of $60,000$ training and $10,000$ testing images. 

\textbf{USPS:} This is also a digit dataset with $7,291$ training and $2,007$ testing images. 

\textbf{SVHN:} Another digit dataset with $73,257$ training, $26,032$ testing, and $53,1131$ extra training images.

\textbf{SYN SIGNS:} This is a synthetic traffic sign dataset, which contains $100,000$ labeled images, and 43 classes.

\textbf{GTSRB:} A dataset for German traffic signs recognition benchmark. The training set contains $39,209$ labeled images and the test set contains $12,630$ images. It also contains 43 classes.

We evaluate the unsupervised domain adaptation model on the following four transfer scenarios:
\begin{itemize}
\item SVHN $\xrightarrow{}$ MNIST 
\item USPS $\xrightarrow{}$ MNIST 
\item MNIST $\xrightarrow{}$USPS 
\item SYNSIG $\xrightarrow{}$ GTSRB
\end{itemize}

 \begin{table}[t]
    \begin{center}
    \begin{tabular}{ | c |c | c | c | c | } 
    \hline
      & \textbf{\footnotesize SVHN} & \textbf{\footnotesize MNIST} & \textbf{\footnotesize USPS} & \textbf{\footnotesize SYNSIG} \\ 
      \textbf{\footnotesize Method} & {\footnotesize to} & {\footnotesize to} & {\footnotesize to} & {\footnotesize to}\\
      & \textbf{\footnotesize MNIST} & \textbf{\footnotesize USPS} & \textbf{\footnotesize MNIST} & \textbf{\footnotesize GTSRB} \\
    \hline
        {\footnotesize Source only} & {\footnotesize 67.1} &{\footnotesize 79.4} &{\footnotesize 63.4} &{\footnotesize 85.1}\\
        \hline
        {\footnotesize DANN} \cite{ganin2015unsupervised}&{\footnotesize 71.1}&{\footnotesize 85.1} &{\footnotesize 73.0$\,\pm\,$0.2} &{\footnotesize 88.7}\\ 
        {\footnotesize ADDA} \cite{tzeng2017adversarial}&{\footnotesize 76.0$\,\pm\,$1.8}& - & {\footnotesize 90.1$\,\pm\,$0.8} & -\\
        {\footnotesize CoGAN \cite{liu2016coupled}}& - & - &{\footnotesize 89.1$\,\pm\,$0.8}& -\\
        {\footnotesize PixelDA} \cite{bousmalis2017unsupervised} & - &{\footnotesize 95.9}& - & -\\
        {\footnotesize ASSC} \cite{haeusser2017associative}&{\footnotesize 95.7$\,\pm\,$1.5}& - & - &{\footnotesize 82.8$\,\pm\,$1.3}\\
        {\footnotesize UNIT} \cite{liu2017unsupervised}&{\footnotesize 90.5} &{\footnotesize 96.0}&{\footnotesize 93.6}& -\\
        {\scriptsize CyCADA \cite{hoffman2018cycada}}&{\footnotesize 90.4$\,\pm\,$0.4}&{\footnotesize 95.6$\,\pm\,$0.2}&{\footnotesize 96.5$\,\pm\,$0.1}& -\\
        {\footnotesize GTA} \cite{sankaranarayanan2018generate} &{\footnotesize 92.4$\,\pm\,$0.9}&{\footnotesize 95.3$\,\pm\,$0.7}&{\footnotesize 90.8$\,\pm\,$1.3}& -\\
        {\scriptsize DeepJDOT \cite{bhushan2018deepjdot}} &{\footnotesize 96.7}&{\footnotesize 95.7}&{\footnotesize 96.4}& -\\
        {\footnotesize SimNet \cite{pinheiro2018unsupervised}}& - &{\footnotesize 96.4}&{\footnotesize 95.6}& -\\
        {\footnotesize GICT} \cite{qin2019generatively}&{\footnotesize 98.7} &{\footnotesize 96.2}&{\footnotesize 96.6}& -\\
                \hline
        {\footnotesize MCD} \cite{saito2018maximum}&{\footnotesize 96.2$\,\pm\,$0.4} &{\footnotesize 96.5$\,\pm\,$0.3} &{\footnotesize 94.1$\,\pm\,$0.3} &{\footnotesize 94.4$\,\pm\,$0.3}\\
         \hline
        \textbf{\small ours (n = 3)} &{\footnotesize 98.2$\,\pm\,$0.1} & \textbf{\footnotesize 98.5$\,\pm\,$0.2} & {\footnotesize 97.0$\,\pm\,$0.1} & {\footnotesize 95.0$\,\pm\,$0.2}\\
        \textbf{\small ours (n = 4)} &{\footnotesize 98.6$\,\pm\,$0.1} & {\footnotesize 98.4$\,\pm\,$0.3} & {\footnotesize \bf 97.1$\,\pm\,$0.1} & {\footnotesize 95.1$\,\pm\,$0.1}\\
        \textbf{\small ours (n = 5)} &{\footnotesize 98.8$\,\pm\,$0.2} & {\footnotesize 98.1$\,\pm\,$0.1} & {\footnotesize 96.1$\,\pm\,$0.3} & {\footnotesize \bf 95.5$\,\pm\,$0.2}\\
        \textbf{\small ours (n = 6)} &\textbf{\footnotesize 98.9$\,\pm\,$0.1} & {\footnotesize 98.0$\,\pm\,$0.1} & {\footnotesize 96.6$\,\pm\,$0.3} & {\footnotesize 95.3$\,\pm\,$0.3}\\
        
    \hline
    \end{tabular}
    \end{center}
    
    \caption{The performance on digit classification and sign classification. The variable $n$ refers to the number of classifiers. We report the mean and the standard deviation of the accuracy obtained over 5 trials.}
    \label{table:1}
\end{table}

{\bf Results:} The experimental results are shown in Table \ref{table:1}. We can clearly see that our proposed method has a significant improvement over the original MCD model with higher accuracy and lower variance. In the SVHN $\xrightarrow{}$ MNIST task, our method has improved the performance by a margin of $2.7\%$. In the MNIST $\xrightarrow{}$USPS task, our method improved by $2.0\%$. In the task of USPS $\xrightarrow{}$ MNIST, our method improved by $3.0\%$. In the SYNSIG $\xrightarrow{}$ GTSRB task, our method improved by $1.1\%$. 

 It can also be seen from Table \ref{table:1} that blindly adding classifiers does not increase the model performance indefinitely. As the number of classifiers increases, the growth rate of the model's performance will decrease and the model's performance will approach a peak range. When the model reaches this peak range, adding more classifiers will cause a decrease in its performance. The most significant increase happens when we increase the number of classifiers from 2 to 3. For example, for USPS-MNIST, the accuracy is increased by 2.0\% when 3 classifiers are employed, however, the performance is only improved by 0.1\% when the number of classifiers is increased from 3 to 4. 

\begin{table*}[htp]
    \centering
    \begin{tabular}{c || c c c c c c c c c c c c || c}
    \hline
       Method & { \small plane} & { \small bcycl} & { \small bus} &{ \small car} &{ \small horse} &{ \small knife} &{ \small mcycl} &{ \small person} &{ \small plant} &{ \small sktbrd} &{ \small train} &{ \small truck} &{ \small mean}  \\
    \hline
       {\small Source Only} & 55.1 & 53.3 & 61.9 & 59.1 & 80.6 & 17.9 & 79.7 & 31.2 & 81.0 & 26.5 & 73.5 & 8.5 & 52.4 \\
       \hline
       {\small MMD} & 72.3 & 53.1 & 64.7 & 31.8 & 58.2 & 14.3 & 80.7 & 60.0 & 70.0 & 41.4 & 89.7 & 20.7 & 55.9\\
       
       {\small DANN \cite{ganin2015unsupervised}} & 81.9 & 77.7 & 82.8 & 44.3 & 81.2 & 29.5 & 65.1 & 28.6 & 51.9 & 54.6 & 82.8 & 7.8 & 57.4\\
       
       
       
       
       
              
       
       
              

                     
       \hline
       {\small MCD \cite{saito2018maximum}} & 87.0 & 60.9 & 83.7 & 64.0 & 88.9 & 79.6 & 84.7 & 76.9 & 88.6 & 40.3 & 83.0 & 25.8 & 71.9\\
       
       
       \hline
       {\small OURS (n = 3)} & 91.1 & 70.0 & 84.1 & 68.8 & 93.5 & 85.3 & 88.2 & 76.7 & 89.2 & 75.0 & 86.2 & 31.2 & 78.3\\
       {\small OURS (n = 4)} & 93.3 & 77.6 & 85.3 & 77.4 & 91.8 & 90.2 & 86.4 & 79.3 & 89.5 & 80.4 & 88.5 & 33.1 & 81.1\\
       {\small OURS (n = 5)} & 95.2 & 81.4 & 82.9 & 82.7 & 91.9 & 89.6 & 86.7 & 79.3 & 91.0 & 66.6 & 88.1 & 35.7 & 80.9\\
       {\small OURS (n = 6)} & 95.0 & 80.1 & 85.5 & 83.1 & 90.5 & 89.9 & 85.9 & 79.2 & 92.8 & 85.3 & 88.3 & 33.9 & 82.5\\

    \hline
    \end{tabular}
    \caption{Results of unsupervised domain adaptation on VisDA2017 \cite{visda2017} image classification task. The accuracy is obtained by fine-tuning
    ResNet-101  \cite{he2016deep} model pre-trained on ImageNet  \cite{imagenet_cvpr09}. This task evaluates the adaptation capability from synthetic CAD model images to real-world
    MS COCO  \cite{lin2014microsoft} images. Our model achieves the best performance in most categories.} 
    \label{table:2}
\end{table*}
  \begin{figure}
    \centering
    \includegraphics[width=1\columnwidth]{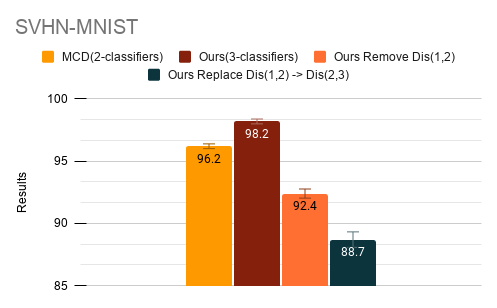}
    \caption{Results of the ablation study}
    \label{fig:5}
\end{figure}

\subsection{Experiments on VisDA Classification Dataset}

 We further evaluate our method on the large VisDA-2017 dataset \cite{visda2017}. 
 The VisDA-2017 image classification is a 12-class domain adaptation dataset used to evaluate the adaptation from synthetic-object to real-object images. 
 The source domain consists of 152,397 synthetic images, where 3D CAD models are rendered from various conditions. 
 The target domain consists of 55,388 real images taken from the MS-COCO dataset  \cite{lin2014microsoft}.
 
 In this experiment, we employ Resnet-101  \cite{he2016deep} as our feature extractor, and the parameters are adopted from the ImageNet pre-trained model. The pre-trained model of our Resnet-18 comes from Pytorch  \cite{paszke2017automatic} and all experimental implementations are based on Pytorch.  
 The input images are of the size $224 \times 224$. First, we resize the input image to 256, and then crop the image to $224 \times 224$ from the center. When we train the model using only the source domain, we just modify the output size of the original last fully connected layer to a size that conforms to VisDA-2017 \cite{visda2017}. In other tasks, we utilize a three-layer fully connected layer structure to replace the one-layer fully connected layer structure of the original classifier. In order to eliminate the interference factors, except for the source only, all other algorithms use the same classifier. We uniformly use SGD as the optimizer for training, and use $ 1\times10^{-3}$ for the learning rate of all methods. We use 16 as the batch size for training.
 
 {\bf Results:} The experimental results of this part are shown in Table \ref{table:2}.
 In this experiment, we achieve similar experimental results with the digits dataset. Compared with the original MCD \cite{lu2020stochastic} method, our proposed multi-classifier implementation method significantly improves the performance in most categories in comparison with the 2-classifier structure. The most prominent increase happens when the number of classifiers is increased from 2 to 3 with the mean accuracy promoted from 71.9\% to 78.3\%.
 

 \subsection{Ablation Study}

 We conducted the ablation study based on the digital datasets (SVHN $\xrightarrow{}$ MNIST) with the same unsupervised domain adaptation setting as Section \ref{IV.B}.
 Since our proposed method can add any number of classifiers, in order to simplify the work, we only use the structure with 3 classifiers as an example in this experiment. We compare our proposed method under three different situations: (i) the original MCD; (ii) arbitrarily remove a pair of discrepancy loss from our Loss; and (iii) arbitrarily replace a discrepancy loss with any existing discrepancy loss.
 The latter two cases simulate the situation where there is a gap in the loss closed loop we proposed. The first one is caused by the missing, and the second is caused by duplication. In this regard, we hope to prove the necessity of the closed structure of the loss function.
 
 The results of the experiment are shown in Figure \ref{fig:5}. It can be seen from the figure that whether it is missing or duplicated, both cases will cause a significant drop in the performance of our proposed method, even lower than the baseline MCD. In the case of repetition, the repeated sub-loss conflicts between the training of the two classifiers, which makes the function difficult to fit, and therefore all classifiers are affected.
 
 \subsection{Convergence and Efficiency Analysis}
 
\begin{figure}[t]
    \centering
    \includegraphics[width=1\columnwidth]{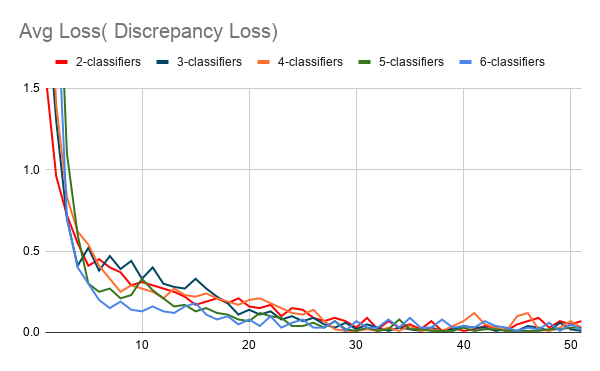}
    \caption{The convergence of different numbers of classifier structures for the USPS-MNIST task. The convergence speed of discrepancy loss will be a little bit faster for more classifiers.}
    \label{fig:1}
\end{figure}

In this section, we make further analysis on the convergence and efficiency of the models. First, we explore the effect of the number of classifiers on the convergence speed of the model using the USPS-MNIST dataset as an example. We record the changes in the model’s average discrepancy loss with iterations and presented them in Fig \ref{fig:1}. From this experiment, we find that the convergence speeds of different models are roughly close to each other. As the number of classifiers increases, the convergence speed of the model will slightly increase.
 
\begin{figure}[t]
    \centering
    \includegraphics[width=1\columnwidth]{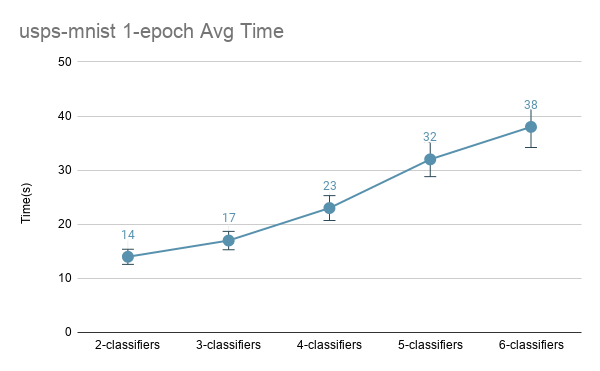}
    \caption{
    The time complexity of our proposed method for the task of USPS-MNIST.
    The experiment is conducted using a RTX 2070 single GPU, and set the batch size to 256, and the other settings are the same as  Section \ref{IV.B}. From the data, we can see that the time complexity increases with the increase of the number of classifiers, however, the time it takes under the structure of three-classifier is close to that of two-classifier.}
    \label{fig:2}
\end{figure}

Second, we compare the average time required to train an epoch with different numbers of classifier structures. We record the training time complexity in the USPS-MNIST experiment and show the result in Fig \ref{fig:2}. From this figure, we can see that, as the number of classifiers increases, the time complexity will increase accordingly. However, the time increase is not too prominent when the number of classifiers is increased from 2 to 3.
 
Based on the above experiments, we can see that the classification accuracy is boosted dramatically for all datasets when we increase the number of classifiers from 2 to 3, while the computation time only increases slightly. Therefore, as a trade-off between performance and efficiency, we believe the structure with 3 classifiers is the best choice in practice for most unsupervised domain adaptation tasks.
 
\section{Conclusion}
 In this paper, we have proposed a straightforward approach of adding classifiers for the adversarial training framework of the maximum classifier discrepancy of the multi-classifier structure. 
 In order to prevent conflicts during the training of multiple classifiers, we have proposed a discrepancy loss function based on the principle that the classifiers are different from each other. This loss function allows us to add any number of classifiers under the original maximum classifier discrepancy framework. 
Extensive experiments show that the multi-classifier structure has distinct performance advantages over the original 2-classifier structure. In all classification tasks, the performance has improved dramatically by adopting the structure with three classifiers.
 





\section*{ACKNOWLEDGMENT}
The work was supported in part by The National Aeronautics and Space Administration (NASA) under grant no. 80NSSC20M0160, and the Natural Sciences and Engineering Research Council of Canada (NSERC) under grant RGPIN-2021-04244.


\balance


\end{document}